# Comprehensively identifying Long Covid articles with human-in-the-loop machine learning

Robert Leaman[1], Rezarta Islamaj[1], Alexis Allot[1], Qingyu Chen[1], W. John Wilbur[1], and Zhiyong Lu[1,2,*]

[1]National Center for Biotechnology Information, National Library of Medicine, National Institutes of Health 8600 Rockville Pike, Bethesda, MD, USA

[2]Lead contact

[*]Correspondence: zhiyong.lu@nih.gov

## Highlights

- We classify COVID-19 articles for relevance to Long Covid, a novel condition
- Most Long Covid articles do not mention it by name, complicating identification
- We ensemble differing data views for robust prediction and to direct human annotation
- We created the Long Covid collection, ~9,000 articles, available in LitCovid

## The bigger picture

Long Covid causes ongoing multisystemic symptoms in a substantial percentage of COVID-19 survivors and lacks specific treatments. Locating articles that refer to novel entities such as Long Covid is generally challenging since keyword searches suffer from limited results and low accuracy without broadly supported terminology. We developed an iterative human-in-the-loop framework to comprehensively identify articles relevant to Long Covid. Our framework integrates multiple classifiers with complementary views and varying accuracy into a single model that reliably predicts the relevance of each article to Long Covid and its priority for manual annotation. We show that most articles relevant to Long Covid do not name the condition and are missed by keyword search. We present and analyze a comprehensive collection of Long Covid articles in LitCovid, which we believe will help accelerate research into this pressing public health issue.

## Summary

A significant percentage of COVID-19 survivors experience ongoing multisystemic symptoms that often affect daily living, a condition known as Long Covid or post-acute-sequelae of SARS-CoV-2 infection. However, identifying scientific articles relevant to Long Covid is challenging since there is no standardized or consensus terminology. We developed an iterative human-in-the-loop machine learning framework combining data programming with active learning into a robust ensemble model, demonstrating higher specificity and considerably higher sensitivity than other methods. Analysis of the Long Covid collection shows that (1) most Long Covid articles do not refer to Long Covid by any name (2) when the condition is named, the name used most frequently in the literature is Long Covid, and (3) Long Covid is associated with disorders in a wide variety of body systems. The Long Covid collection is updated weekly and is searchable online at the LitCovid portal:
https://www.ncbi.nlm.nih.gov/research/coronavirus/docsum?filters=e_condition.LongCovid

## Graphical Abstract

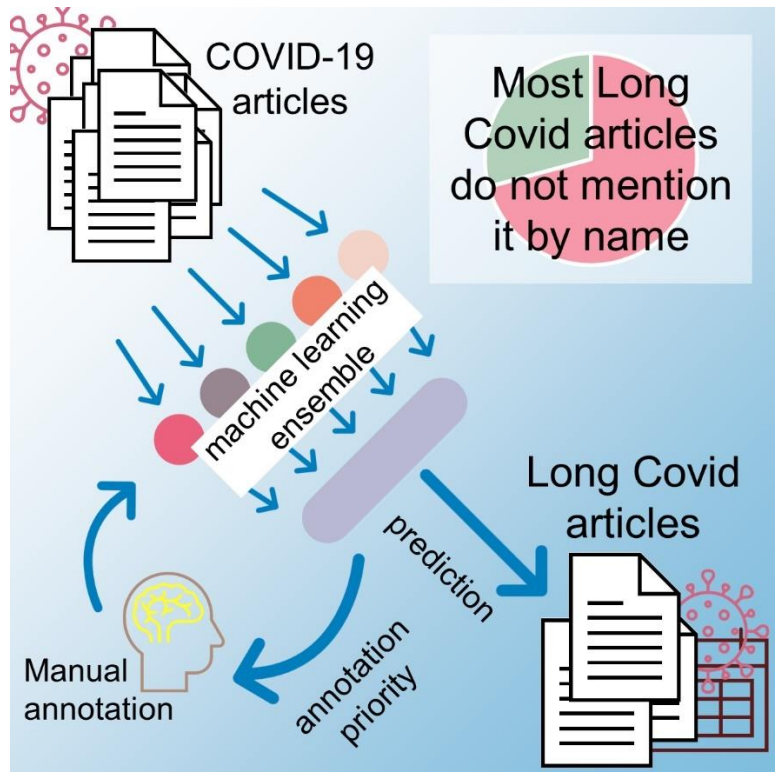

## Data Science Maturity

DSML 2: Proof of concept: Data science output has been formulated, implemented, and tested for one domain/problem.

## Introduction

Literature collections such as LitCovid provide a critical resource as scientific understanding expands, serving as a centralized access point for reliable and comprehensive information on COVID-19.[1] LitCovid initially launched in February 2020, providing a set of eight topics, such as prevention and diagnosis, to improve information accessibility.[2] As our understanding of the effects of COVID-19 continues to evolve, however, updates are necessary.[3] In this work we identify articles which discuss the long-term complications of COVID-19.

Early in the COVID-19 pandemic, some COVID-19 patients began reporting symptoms persisting significantly past the acute phase. Finding existing supports lacking, these patients – many of whom were themselves healthcare professionals or researchers – turned online for support, naming the condition Long Covid, as a contraction of long-term COVID illness.[4] In May 2020, a patient-led group published the first survey of long-term symptoms of COVID-19.[5] Extensive subsequent research continues to show that a significant percentage of COVID-19 survivors experience ongoing multisystemic symptoms.[6-8] These symptoms include respiratory issues, cardiovascular disease, cognitive impairment and profound fatigue.[9-12] For many patients, these symptoms affect daily living or returning to work.[13,14] Long Covid occurs in patients with low risk of fatal outcome and in younger patients, including children.[13,15,16] Most of the morbidity burden of COVID-19 (i.e., healthy years of life lost) is in COVID-19 survivors, not fatalities.[17] Moreover, many viruses besides SARS-CoV-2 – including Poliovirus, Varicella-Zoster, Epstein-Barr, Zika, West Nile, and SARS-CoV – have been implicated in long-term sequelae.[18-23]

Long Covid remains incompletely understood, however, despite increasing evidence for several theories and notable overlaps with other conditions, including myalgic encephalomyelitis/chronic fatigue syndrome.[24,25] Reported incidence rates vary widely, from 9% to 81% according to one meta-analysis.[26] Evidence for widely effective treatments is lacking.[27] While consensus-based case definitions are emerging, definitions of Long Covid used in the literature vary substantially, which impairs building on previous work.[28-30] Nevertheless, there is increasing recognition that COVID-19 is not only a mass death event, but – through Long Covid – also a mass disabling event, making it a pressing public health concern.

Our initial analysis of the published literature found a wide variety of terms used to refer to Long Covid, but also found that the condition is more commonly described rather than named. Querying for Long Covid articles is therefore challenging: precise queries such as "post-acute sequelae of SARS-CoV-2 infection" return limited results while broad queries such as "post COVID symptoms" return many false positives.

In this work, our goal is to identify biomedical research articles relevant to Long Covid which are useful to researchers, clinicians, and patients/advocates. Theoretically, the task is binary text classification task.[31,32] However, the objective is to comprehensively identify uncommon articles describing a novel disease entity – Long Covid – that is incompletely understood, inconsistently defined, lacks established terminology, and is frequently not named. The class imbalance and large number of articles to be classified suggest actively choosing which articles to annotate manually. We therefore employ a human-in-the-loop approach utilizing active learning to identify the articles where manual annotation would be most useful.[33] In preliminary experiments, however, we found conventional methods – uncertainty sampling with either classical machine learning or transformer-based deep learning – failed to differentiate relevant articles with language differences from the large number of irrelevant articles.

Our work therefore emphasizes thorough data exploration. We utilize multiple relevance signals as differing views of the data to identify areas of disagreement between individual signals. We also break the manually annotated data reserved for training into multiple subsets, training models on each to identify articles whose predictions are not based on robust patterns. We further utilize sources of labels available without training data, which are sometimes noisy. We combine these approaches using the weakly supervised method data programming, which integrates a set of task-specific noisy signals, called labeling functions, without additional training data.[34]

The contributions of this article are three-fold. First, we report the creation of the Long Covid collection, a literature resource of 8,950 articles (through July 29, 2022) relevant to an urgent public health concern. The Long Covid Collection is publicly available within the LitCovid portal, a widely used literature hub with over 290,000 articles specific to COVID-19. Second, we present an analysis of the Long Covid collection, demonstrating that 69.0% of relevant articles do not mention Long Covid directly, making identification via query difficult. Third, we present a framework for comprehensively identifying articles relevant to concepts without established terminology, combining human-in-the-loop machine learning and data programming. We further present three extensions to data programming. We evaluate the automated prediction model on a held-out set of manually annotated articles, demonstrating an ROC AUC of 0.8454. We also compare our approach to several other approaches to identifying Long Covid articles, demonstrating an over 3-fold improvement in sensitivity.

## Results

### Definition and Guidelines

Following the broadest early definition with substantial support, we define Long Covid to be ongoing symptoms at least four weeks after initial symptoms.[35] We therefore label an article as relevant to Long Covid if it meets the following two criteria: First, the article must consider adverse effects resulting from COVID-19, i.e., SARS-CoV-2 infection. Second, the article must report outcomes or symptoms over a timeframe that includes at least four weeks post-infection. While the goal is to label articles as relevant if they contain useful discussion of the long-term symptoms caused by COVID-19, several aspects make these relevance judgements difficult. First, articles do not need to mention Long Covid by name to be relevant. Second, the symptoms may be of any type, provided they are persistent and caused by COVID-19. Third, the relevant discussion of persistent symptoms may occur in the full text rather than the abstract. Finally, articles that only refer to Long Covid in passing – such as to mention that long-term sequelae should be studied – are not relevant, even if they would be returned by keyword search. We provide the full annotation guidelines in the Supplementary Materials.

We performed a small manual inter-annotator agreement study to verify the repeatability of the annotator guidelines. We randomly selected 100 of the articles previously annotated by the primary annotator (RL, a bioinformatician with previous annotation experience). These were labeled by the senior annotator (JW, an M.D./Ph.D.). Each article was labeled as relevant or not relevant, using the full text of the article as needed. The annotators agreed on 87 articles (61 Relevant, 26 Irrelevant) for a raw inter-annotator agreement of 87.0%. Cohen's kappa, which controls for chance agreement and ranges from -1.0 to 1.0, was 0.70, corresponding to substantial agreement.[36]

Nearly all annotator disagreements were due to the difficulty of clearly establishing the timing of the symptoms described, which often requires careful analysis. For example, the timeline for the study described in the full text for PMID 32548209 is clearly not long enough to meet the 4 weeks required by our guidelines. While this would suggest that the article is not relevant, the article does not specify the length of time from initial infection to enrollment in the study and refers to symptoms persisting after clinical recovery. Other annotator disagreements were also primarily due to textual ambiguities; for example, PMID 33756229 refers to patients that test positive again without clearly specifying whether the cases were due to either reactivation – which would be relevant – or reinfection – which would not be relevant.

### Data Summary

Our input dataset consists of LitCovid, a literature resource of COVID-19 articles in PubMed, updated daily. LitCovid categorizes each article with eight broad topics (general information, mechanism, transmission, diagnosis, treatment, prevention, case report, and epidemic forecasting). LitCovid considers all PubMed articles other than preprints. We therefore use LitCovid as a comprehensive collection of articles about COVID-19, and all articles in LitCovid were considered. The oldest articles in LitCovid were published in January 2020.

We created the Long Covid Collection using a human-in-the-loop machine learning process with the goal of minimizing human effort while creating a classifier that is both accurate and able to identify articles requiring human labels due to uncertainty. In our usage, an active learning process iteratively chooses articles for the human annotators to judge for relevance, which are then used to improve an automated system. The updated system is then used to select a new set of articles for annotation, focusing on articles where the automated system is uncertain, and the process repeats. The automated system is designed as an ensemble of lightweight, independent, classifiers with differing views of the data. This design is critical for focusing the human annotation effort on articles where the model is uncertain. Our iterative process provides two features not available with a more conventional approach: first, it produces a high probability of identifying all articles relevant to Long Covid, and second, it uses human annotation effort more efficiently.

The Long Covid collection was first released on August 1, 2021, consisting of 2,056 articles, and is updated weekly. As of July 29, 2022, the Long Covid collection contained 8,950 articles, gaining approximately 133 articles per week on average. Approximately 2.9% of articles in LitCovid are relevant, substantially skewing the relevant and irrelevant classes. Our annotation process prioritizes articles where the automated system is uncertain, and this high skew results in the least certain articles containing a high proportion of relevant articles. As a result, the manually annotated articles contain a greater number of relevant articles than irrelevant articles. Moreover, the irrelevant articles manually annotated tend to be those that are difficult to distinguish from relevant articles. As of July 29, 2022, there were 10,149 manually annotated articles, 5,800 annotated as relevant and 4,349 annotated as irrelevant. As new articles are annotated manually, one quarter are randomly reserved for validation.

### Validation: Comparison Methods and Evaluation

We compared our results to several other collections on Long Covid. The CoronaCentral resource contains articles related to several coronaviruses, including SARS-CoV-2, with automated predictions for both topics and various entities.[37] We consider articles annotated with both SARS-CoV-2 and the Long Haul topic as Long Covid articles according to CoronaCentral. PubMed Clinical Queries uses predefined keyword filters to help users perform and refine specialized searches.[38] The queries for COVID-19 are intended to limit results to articles on SARS-CoV-2 with a particular topic; we use the Long COVID filter, which is implemented as a keyword query and listed in full in the Supplementary Materials. Medical Subject Headings (MeSH) is a controlled vocabulary used for indexing articles by topic.[39] We created a Long Covid query from MeSH terms by combining COVID-19 or SARS-CoV-2 with terms reflecting the post-acute phase, also listed in full in the Supplementary Materials. Finally, we created several textual queries from the most common Long Covid terms. These queries are: post covid symptoms, "Long Covid," "post-acute sequelae" AND ("SARS-CoV-2" OR "COVID-19"), and "post covid syndrome."

The evaluation set, created by randomly reserving one quarter of all articles annotated manually, contains 1,450 positive articles and 1,088 negative articles. We evaluate the results using sensitivity and specificity, which can be visualized using the Receiver Operating Characteristic (ROC) curve, and can be summarized as the area under the ROC curve (AUC).[40] Since the comparison approaches provide only binary predictions, we binarize our results for comparison by thresholding at a prediction of 0.7. The ROC curve for our results and the sensitivity/specificity points for all comparison approaches can be seen in Figure 1. The area under the curve (AUC) is 0.8454.

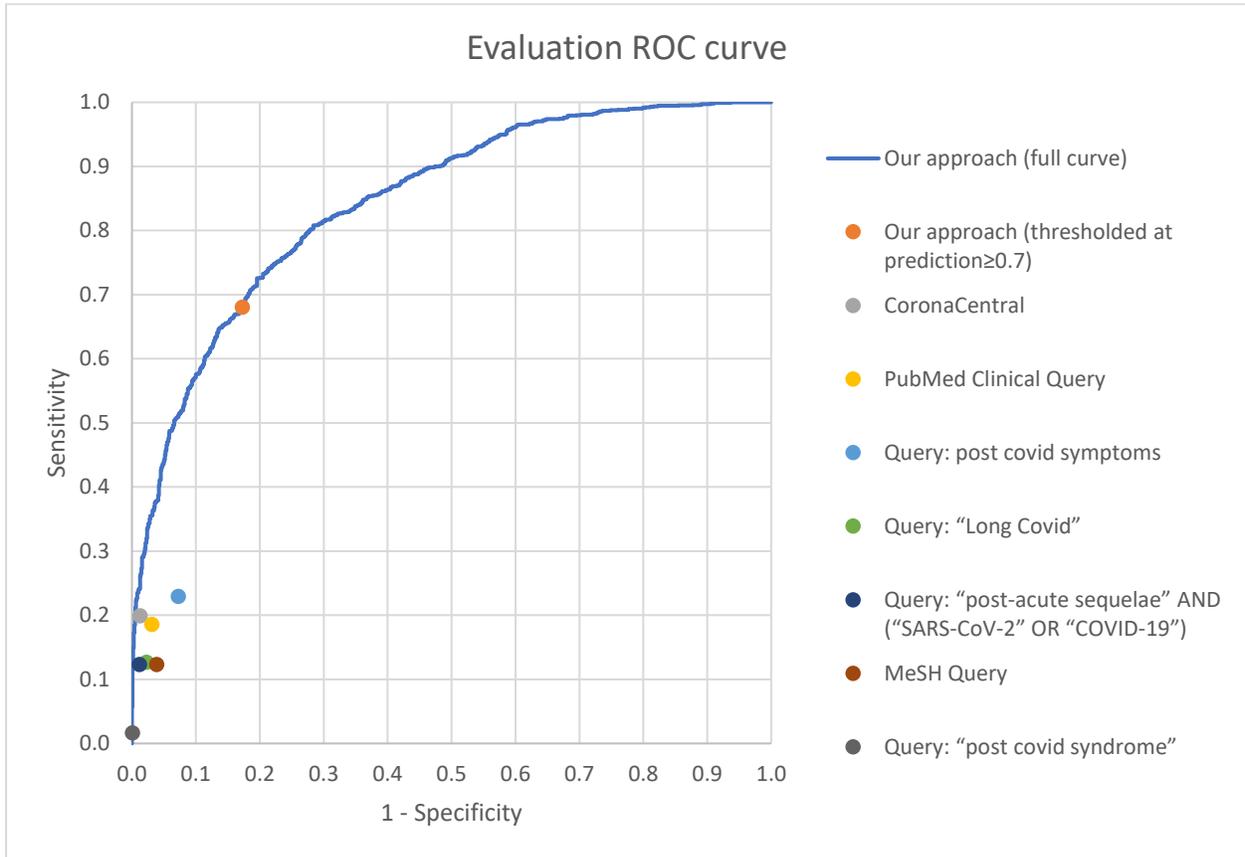

*Figure 1.* Receiver Operating Characteristic (ROC) curve of our results. Shown with the sensitivity/specificity points for our results thresholded at prediction ≥ 0.7 and several alternative methods of collecting articles relevant to Long Covid. The area under the curve (AUC) is 0.8454.

While the comparison methods provide high specificity, the highest sensitivities are for the post-covid symptoms query (0.2297) and CoronaCentral (0.1993), while the sensitivity of our thresholded results is 0.6807.

### Resource Analysis

We created a grammar-based named entity recognizer to identify mentions of Long Covid. This grammar is designed to accommodate significant language variability and extends a grammar previously created to identify mentions of COVID-19 and SARS-CoV-2.[41] The Long Covid grammar-based named entity recognizer identified 7,378 mentions of Long Covid, representing 763 unique phrases (after normalizing case and punctuation); the most frequent are summarized in Figure 2.

Despite the flexibility of the grammar, 69.0% of the articles in the Long Covid Collection do not contain an identifiable term for Long Covid. This is commonly caused by the article referring to Long Covid using a description rather than a term. While grammar can identify many descriptive phrases, such as "long-term outcomes of COVID-19," some of the descriptions used by authors to refer to Long Covid remain beyond the ability of the grammar to recognize. This is often due to some qualification, such as an anatomical system. For example, the phrase "residual respiratory impairment after COVID-19 pneumonia" [PMID 34273962] strongly suggests a respiratory form of Long Covid but could not be recognized by the grammar. A more advanced recognition technique should be able to recover some additional descriptive mentions, however any such mentions remaining are not common. A more advanced technique may also help reduce false positives in the grammar, though these are quite rare. For example, "… how long COVID-19 (SARS-CoV-2) survives …" [PMID 32967479] includes the phrase "long COVID" but does not refer to Long Covid.

Inspection of the term frequencies, as seen in Figure 2 and in the full data, show that the frequency of Long Covid terms reflect a long tail distribution. Plotting the rank of each term against its frequency in a log-log plot results in an approximately straight line (data not shown), indicating a Zipf distribution, as is common for linguistic data.[42] Identifying Long Covid articles by identifying synonymous terms is therefore subject to diminishing returns and additional methods are required.

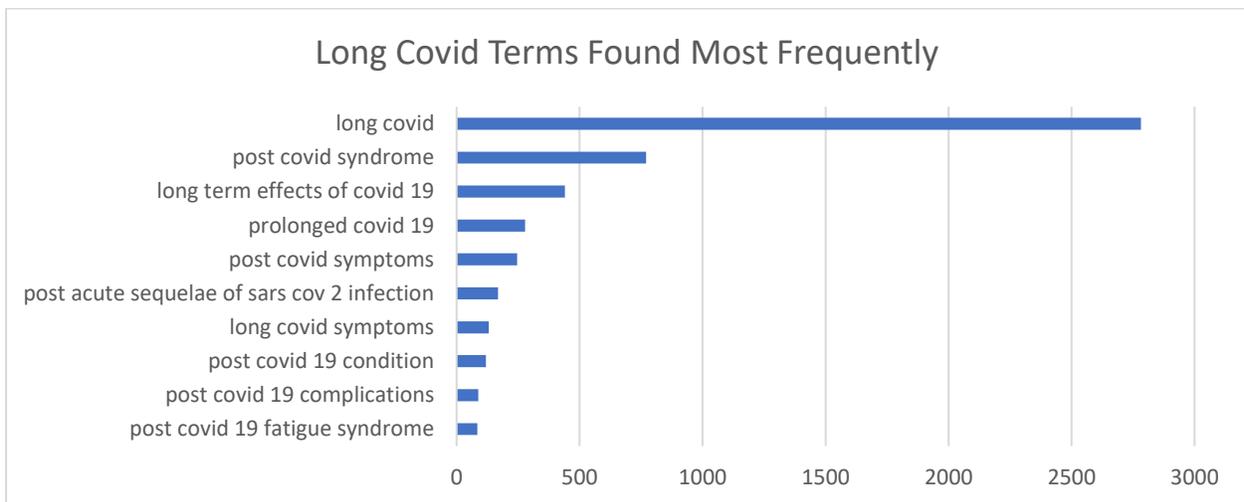

*Figure 2.* Terms for Long Covid found most frequently by the Long Covid grammar. The grammar found a total of 7,378 mentions of Long Covid, representing 763 unique phrases, ignoring capitalization and punctuation.

### Naming Trend for Long Covid over Time

Figure 3 shows the naming trend over time, with all articles relevant to Long Covid listed as either mentioning Long Covid directly (i.e. using the term Long Covid), mentioning Long Covid but using a different term, or not mentioning Long Covid by name. All articles listed are relevant to Long Covid, and each article is only counted once.

We see that not only is Long Covid the most common term used in the literature to refer to Long Covid (see Figure 2), but its use also appears to be increasing slowly. Moreover, the percentage of articles which are relevant to Long Covid but do not refer to it using an identifiable term appears to be decreasing. However, these changes appear to be gradual, suggesting that the lack of terminological consensus will remain for some time. Unfortunately, this reluctance to name the condition likely makes it more difficult for consensus to build: articles that rely on descriptions will be more difficult to locate since automated recognition of descriptions is known to be much more difficult than names.[43]

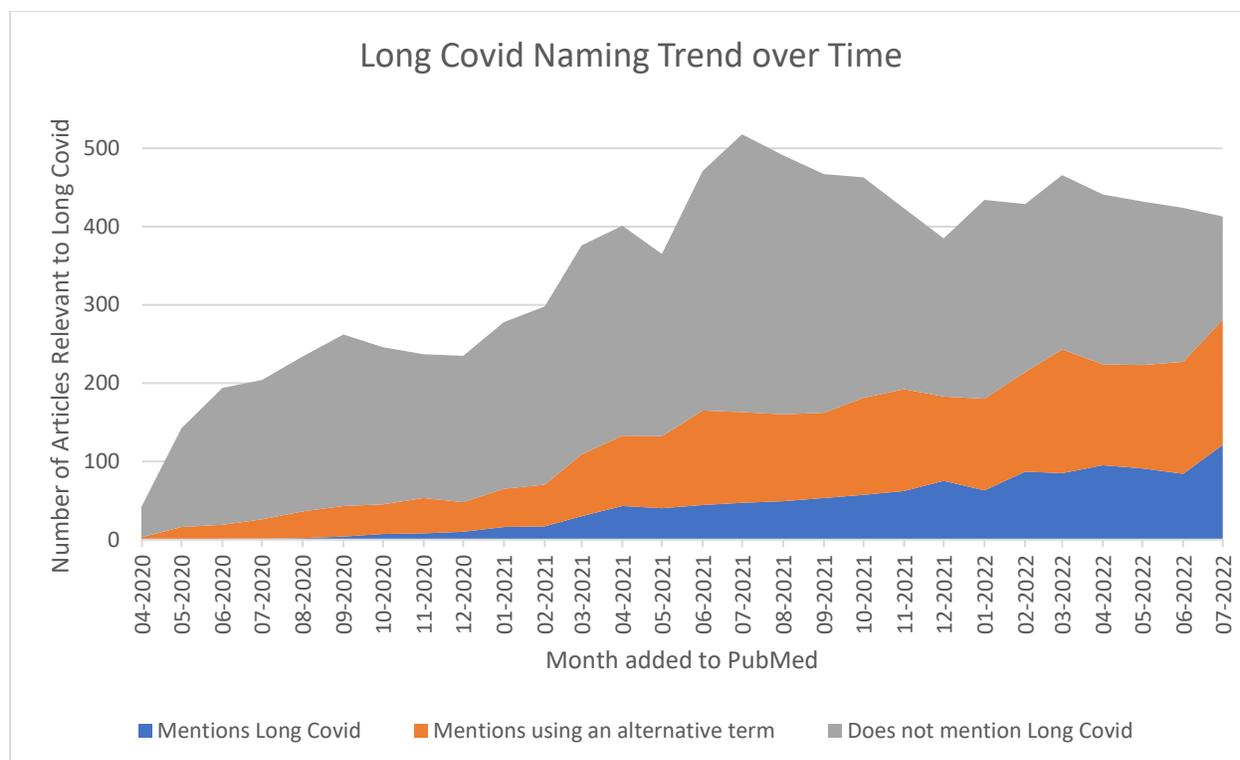

*Figure 3*. Trend over time of terms used to refer to Long Covid in articles relevant to Long Covid. Articles that mention Long Covid use the name Long Covid at least once. Articles that use an alternative term mention Long Covid at least once via identifiable synonym. Articles that do not mention Long Covid do not contain an identifiable term for Long Covid. All articles listed are relevant to Long Covid.

### Analysis of Entities Mentioned

PubTator is a web-based system providing annotations for six entity types: genes/proteins, genetic variants, diseases, chemicals, species, and cell lines.[44] We compared the annotation rate for entities annotated by PubTator in the Long Covid Collection and LitCovid. We found that the entities which showed a statistically significant difference ($p < 0.01$, Fisher's exact test) were primarily disorders that appear more frequently in the Long Covid Collection than in the general COVID-19 literature. We selected for these and further removed disorders specific to COVID-19, such as multisystem inflammatory syndrome in children (MIS-C). Figure 4 visualizes the 30 disorders that appear most frequently in the Long Covid collection in a dendrogram, clustered according to the number of ancestors in common in the MeSH hierarchy.

Figure 4 underscores the great variety of body systems affected by Long Covid, demonstrating that Long Covid is a multisystemic condition. Several of the symptoms most common in Long Covid patients are listed, such as fatigue and cognitive dysfunction.[6] Neurological and cardiovascular conditions also appear prominently. A notable trend which is less apparent in this clustering are conditions due to immune system dysregulation, such as Guillain-Barre syndrome, myocarditis (inflammation of the heart muscle), encephalitis (inflammation of the brain) and inflammation itself. Interestingly, several symptoms closely associated with COVID-19 are seen to appear more frequently in connection with Long Covid than COVID-19, such as respiratory symptoms, dyspnea, and olfaction disorders.

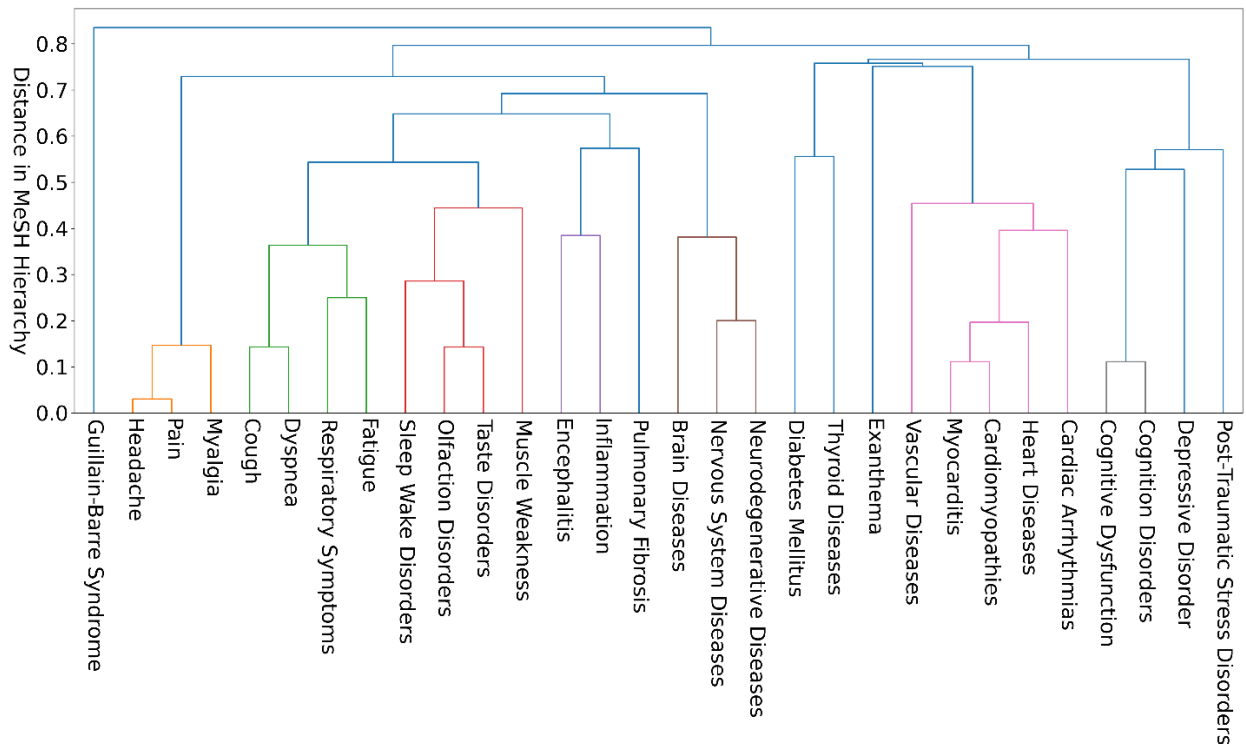

*Figure 4.* Dendrogram of the disorders that appear most frequently in the Long Covid collection and whose annotation rate is greater than in the general COVID-19 literature (p < 0.01, Fisher's exact test). Disorders are clustered according to the number of ancestors in common in the Medical Subject Headings (MeSH) hierarchy.

The chemicals which appear statistically significantly more frequently in the Long Covid collection than in LitCovid include several classes of drugs. These are steroids (prednisolone, methylprednisolone), NSAIDs (aspirin, indomethacin), and anti-fungals (amphotericin B). Other chemicals that appear more frequently in the Long Covid collection are chemicals used in various clinical tests (gadolinium, fluorodeoxyglucose F18, carbon monoxide), and cortisone, due to its relation to adrenal insufficiency.

## Analysis of Topic Clusters

We use the probabilistic distributional clustering (PDC) algorithm to identify topics within the Long Covid Collection.[45] PDC uses terms, phrases, and MeSH terms occurring within a collection as input and utilizes their probability of co-occurrence to partition the set of input features into disjoint groups. Documents can then be scored with respect to each topic identified and may receive a high score for more than one topic. Figure 5 shows the most frequent topics identified by the PDC clustering algorithm, organized into four aspects: study methods, interventions (treatments and tests), systemic dysfunctions and specific disorders. Articles which are identified as containing multiple topics within a chart contribute fractionally to each topic, so that each article is counted only once. The names for each topic are manually generated but represent the most common phrases in the topic.

Broadly, the topics show an expansion through mid-2021, with a slight contraction after. Figure 5a shows significant expansions for both cohort studies and systematic reviews, while case studies show a slight contraction since late 2021. This pattern suggests increasing scientific rigor over time.[46] Interventions, in Figure 5b, include both testing and treatments but were not common. Moreover, the interventions which are seen are not specific to Long Covid, but rather broad care classifications or extensions of COVID-19 interventions. Figure 5c describes systemic dysfunctions; we see that the initial discourse was dominated by pulmonary dysfunction, though discussion of neurological & cognitive dysfunction topic also began early. All other systemic dysfunction topics start small but increase over time, indicating increasing recognition of the long-term effects of COVID-19 on multiple body systems. Specific disorders appear in Figure 5d. These disorders affect a wide variety of body systems, and – with one clear exception – all follow the general trend of increasing counts. The exception, the viral pneumonia topic, is the only specific disorder topic that starts with considerable counts, then contracts significantly at the end of 2020. This trend shows that acute COVID-19, which primarily causes viral pneumonia, is typically no longer discussed with Long Covid: Long Covid is now discussed as a separate entity.

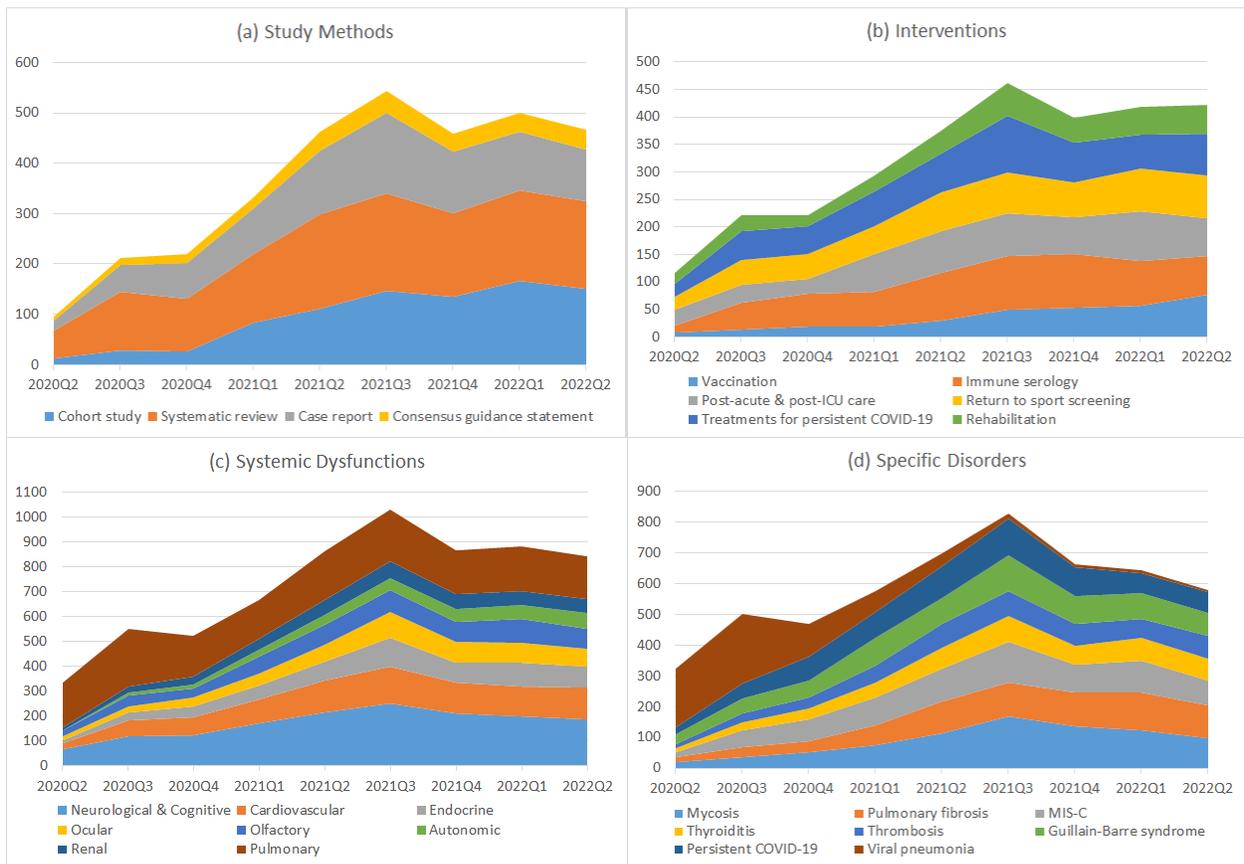

*Figure 5.* Most frequent topics in the Long Covid collection over time. Topic names are manually generated but reflect the most common phrases in the topic. Study types (a) show increased rigor over time, while Interventions (b) show a lack of treatments or tests specific to Long Covid. Long Covid is a complex, multisystemic, condition that causes a wide variety of potentially serious dysfunctions (c) and disorders (d).

## Discussion

COVID-19 has caused widespread mortality and strained healthcare systems worldwide.[47] Estimates of the overall morbidity burden show, however, that most of the burden lies in COVID-19 survivors.[17] While estimates of the prevalence of Long Covid vary, a recent meta-analysis shows that the prevalence of symptoms beyond four weeks is quite high: 43%.[26] Moreover, the effects experienced many years or decades later are yet unknown. Continuing research into Long Covid is critical and identifying Long Covid articles comprehensively allows the articles to be analyzed as a set. Our analyses of both the entity mentions and the topic clustering support the view that Long Covid is a multisystemic condition. Notably, our experiments did not uncover support for psychological causes. However, both the analysis of entity mentions and the topic clustering find a greater number of symptoms associated with Long Covid than treatments for those symptoms.

We also find that there is a significant overlap between the symptoms associated with COVID-19 and the symptoms associated with Long Covid. This supports the view that the name Long Covid is descriptive of the condition as extended COVID. The fact that the name Long Covid is both the most common name and an increasing number of articles use the name suggests that a consensus is building, albeit slowly. However, this does not preclude another name – or subtype name – gaining favor once the etiology – or etiologies – are identified. Interestingly, other names were initially used for both COVID-19 and SARS-CoV-2, pneumonia of unknown aetiology and 2019-nCoV, respectively, though a consensus built fairly quickly.[41] Unlike COVID-19 and SARS-CoV-2, however, standardizing organizations have not argued for specific terminology for Long Covid; use of the term Long Covid is largely due to patient advocacy efforts.[4]

Our definition of Long Covid is primarily time-based, while many of the studies in the literature that discuss sequelae of COVID-19 are primarily concerned with a specific body system. This is particularly the case with neurological effects, which are prevalent in Long Covid but may also appear much earlier, even as the initial manifestation of infection. Our annotators noted that many articles could not be labeled from just the title and abstract; this is primarily because of the condition that symptoms must be present at least a month after initial infection. Moreover, our combination of a broad definition and short timeframe (one-month post-infection) implies the inclusion of some post-COVID conditions with an acute presentation, notably multisystem inflammatory syndrome in children (MIS-C) and mucormycosis. However, these conditions do share some elements, such as immune system dysregulation. Again, Long Covid remains an area of active research and updates are expected.

Our work has several limitations. Our analyses are mostly correlational, and do not show, for example, that the association between specific symptoms are definitively caused by Long Covid. Our method does not specifically address language drift, though the effects of language drift should be ameliorated somewhat through the iterative annotation process. We anticipate that the Long Covid collection itself will remain relevant for some time even if a strong consensus builds and the level of terminological variation drops significantly. Nevertheless, Long Covid remains an area of active research, and new developments are expected.

Our methods are automated and do not produce perfect accuracy. However, this is partially due to inherent ambiguities. For example, it is sometimes difficult to label articles based on the title and abstract, such as determining whether a reference to COVID-19 patients refers to patients who had COVID-19 previously or patients who currently have acute COVID-19 [e.g. PMID 35043098]. Unfortunately, the full text is often not available.

While it is difficult to provide a definitive discussion of how many articles must be annotated to provide high coverage, we note that our framework already provides an AUC of 0.7079 using only the labeling functions that do not require training. We performed a series of experiments iteratively rerunning article selection and found that 1,000 manually annotated articles reliably produces a model with an AUC of over 0.80 (data not shown). In this work we also intended to provide a reliable, comprehensive collection of articles on Long Covid; we therefore proceeded to manually annotate most of the relevant articles and a nearly equal number of irrelevant articles. Note, however, that this still results in a significant annotation savings compared with manual annotation: we annotated approximately 3.7% of the articles in LitCovid, representing an annotations savings of 96.3%. We therefore believe that our framework reduces the need for manual annotations when identifying high-variation terminology.

# Experimental Procedures

## Lead contact

Further information and requests for resources should be directed to the lead contact, Zhiyong Lu (zhiyong.lu@nih.gov).

## Materials availability

This study did not generate any new unique materials.

## Datasets and Software

Software is implemented using Python 3.8, with the following modules: bioc 1.3.5 for document handling; eutils 0.6.0 for querying NCBI e-utilities for PubMed article metadata; pyparsing 2.4.7 for building the Long Covid grammar; scikit-learn 1.0.2, numpy 1.18.5, and scipy 1.8.0 for numerical and statistical methods. Initial input is the LitCovid resource, which contains all published PubMed articles relevant to SARS-CoV-2 or COVID-19 and is updated daily.[1] LitCovid data is downloaded via the public API at https://www.ncbi.nlm.nih.gov/research/coronavirus-api/export/all/tsv. CoronaCentral data is downloaded from Zenodo and updated weekly; version 84 is used for validation.[37,48] PubTator data is downloaded via the public API at https://www.ncbi.nlm.nih.gov/research/pubtator/api.html. Article annotation is performed using the LitSuggest online interface, https://www.ncbi.nlm.nih.gov/research/litsuggest/. The query used to initialize the iterative annotation process was "long covid" OR ("sequelae" AND ("COVID-19" OR "SARS-CoV-2")).

## Methods

### Human-in-the-Loop Process Overview

Our goal is to comprehensively identify all PubMed articles relevant to Long Covid. These articles are a subset of articles relevant to COVID-19, which are already collected by the LitCovid resource. The task therefore becomes classifying each article in LitCovid as relevant or irrelevant to Long Covid. However, since the objective is to identify articles relevant to a novel disease entity that is incompletely understood, inconsistently defined, lacks established terminology, and is frequently not named, our approach prioritizes data exploration in addition to prediction.

In human-in-the-loop machine learning, data points can be selected for human annotation based on either diversity sampling or uncertainty sampling.[33] In diversity sampling, the data is clustered and instances are chosen to ensure that each cluster is represented. Uncertainty sampling, on the other hand, prioritizes instances closest to the decision boundary or instances with the largest variation in predictions. Our initial analysis showed fast improvements with uncertainty sampling, with no additional benefit with diversity sampling. Our framework therefore prioritizes articles for annotation when close to the decision boundary or when the automated predictions show high variation. In preliminary experiments, we found conventional methods – uncertainty sampling with either classical machine learning or transformer-based deep learning – failed to differentiate relevant articles with language differences from the large number of irrelevant articles.

The automated system therefore employs a semi-supervised approach, allowing predictions from multiple relevance signals. The system combines these signals – some of which are created with supervised classification – using the weakly supervised framework data programming.[34] These relevance signals, called labeling functions, are derived from disparate data sources, producing multiple views of the data which are sometimes contradictory. Data programming uses the labeling functions to create an ensemble model without training data.[34] Triplet methods are a recent development in data programming which provide a closed form solution that does not require iterative training.[49] We extend triplet data programming to allow probabilistic labeling functions (rather than only binary), to improve reliability of the ensemble model, and to provide uncertainty scores.

We provide an overview of our framework in Figure 6, which illustrates the data flow for the three high level processes used by our system. The first process, model creation, prepares the labeling functions (some of which require training data), retrieves a label for each article from each labeling function, and creates the ensemble model using data programming. The second process, article prediction, uses the model to predict the relevance of every article. The third process, article annotation, uses the model to identify uncertain predictions; articles with high uncertainty are then prioritized for manual annotation. One quarter of annotated articles are reserved for evaluation and the remainder are added to the training data.

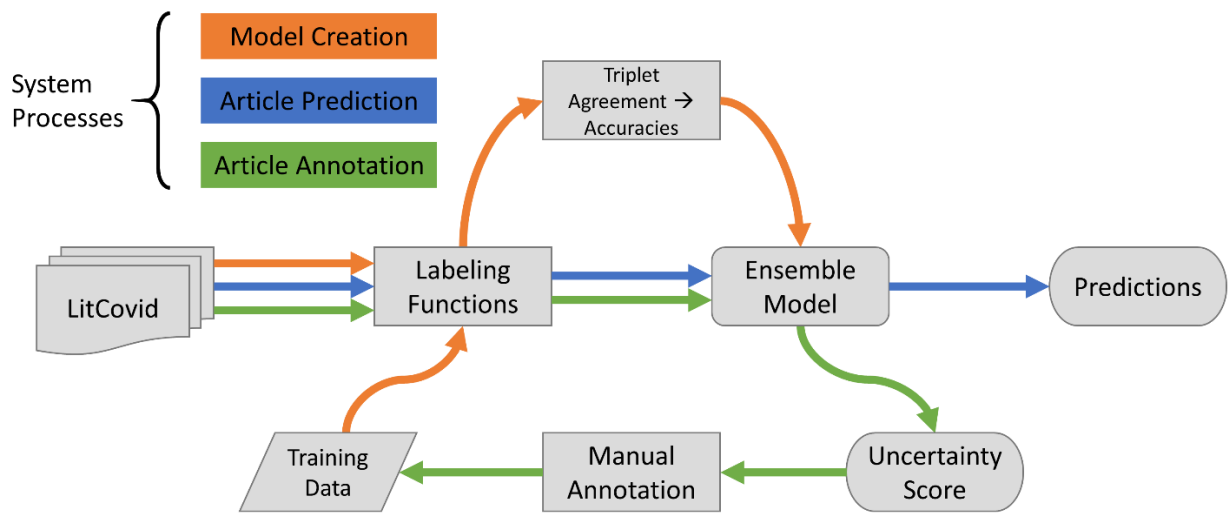

*Figure 6*. System overview, illustrating the flow of data for the three primary system processes: model creation, article prediction and article annotation.

## Data Programming

While supervised machine learning requires abundant training data, data programming creates a model by aggregating weaker forms of supervision.[34] This creates an ensemble model, similar to the machine learning method stacked generalization, also known as stacking.[50] In the data programming paradigm, the practitioner creates labeling functions – task-specific functions that imperfectly label instances – instead of labeling instances. Labeling functions may take many forms, including rule-based patterns, dictionary lookups, and noisy supervised classifiers. The labeling functions are applied to a large amount of unlabeled data, and the agreement rates between the labeling functions are then used to infer the accuracy of each labeling function. The accuracies for each labeling function then form the parameters of a generative model that can be used to label new data points, applying a small amount of knowledge – in the form of labeling functions – to accurately label a large amount of data. Note that various forms of noise – including missing values and disagreements – are anticipated and handled by the framework.

Since human-in-the-loop methods begin with very little labeled data, applying data programming within a human-in-the-loop approach would seem to be ideal. However, the human-in-the-loop approach requires the automated system to be repeatedly retrained, which is inconvenient despite the simplicity of the generative model due to the large data sizes involved. Data programming with triplet methods makes repeated retraining unnecessary by directly estimating the accuracy of each labeling function using a closed form solution.[49] This solution requires only calculating the pairwise agreement rates between a triplet of labeling functions; it can be extended to an arbitrary number of labeling functions by iterating through all possible triplets and averaging the results.

Specifically, given a triplet of binary labeling functions $\mathcal{L}_1$, $\mathcal{L}_2$, and $\mathcal{L}_3$ of the form $\mathcal{L}(x) \in \{0, 1\}$ which are conditionally independent given the class, the estimated accuracy of $\mathcal{L}_1$ is:

$$estimated\_accuracy(\mathcal{L}_1) = \frac{1}{2}\sqrt{\frac{agree(\mathcal{L}_1, \mathcal{L}_2) \cdot agree(\mathcal{L}_1, \mathcal{L}_3)}{agree(\mathcal{L}_2, \mathcal{L}_3)}} + \frac{1}{2}$$

Where we define *agree* for a pair of labeling functions as:

$$agree(\mathcal{L}_1, \mathcal{L}_2) = \frac{1}{|X|} \sum_{x \in X} equal(\mathcal{L}_1(x), \mathcal{L}_2(x))$$

And we define *equal* as:

$$equal(\mathcal{L}_1(x), \mathcal{L}_2(x)) = \begin{cases} 1, if\ \mathcal{L}_1(x) = \mathcal{L}_2(x) \\ -1, if\ \mathcal{L}_1(x) \neq \mathcal{L}_2(x) \end{cases}$$

Note that in the previous definition the labeling functions $\mathcal{L}(x)$ have binary values. We extend the accuracy calculation to handle probabilistic labeling functions $\hat{\mathcal{L}}(x) \in [0, 1]$. We first define $sample(x)$ to be a function that discretizes its input $x \in [0, 1]$ by returning the value 1 with probability $x$ and returning 0 with probability $1 - x$. We can then define the *equal* function for probabilistic labeling functions by averaging over many samples, each of which has discrete binary values. We note that the result converges to a closed form solution as the number of samples approaches infinite, specifically:

$$equal\left(\hat{\mathcal{L}}_1(x), \hat{\mathcal{L}}_2(x)\right) = \lim_{n \to \infty} \frac{1}{n} \sum_n equal(sample(\hat{\mathcal{L}}_1(x)), sample(\hat{\mathcal{L}}_2(x)))$$
$$= (2\hat{\mathcal{L}}_1(x) - 1) \cdot (2\hat{\mathcal{L}}_2(x) - 1)$$

Also note that under this formulation, a labeling function that cannot provide a prediction for any given input may abstain by returning exactly 0.5.

The accuracy of each labeling function can be estimated by forming a triplet with any other two labeling functions. Since there is an abundance of labeling functions, we gather many estimates for each labeling function by using all available pairs and use their mean as the final accuracy estimate. However, we improve the reliability of the accuracy estimates by ignoring pairs of labeling functions whose agreement may be due to chance. We model the agreement between a pair of labeling functions as a binomial distribution, where the number of trials is the number of instances and the number of successes is the number of instances where the labeling functions agree. We calculate a confidence interval on the agreement rate between each pair of labeling functions and discard the pairs where the confidence interval includes 0.5. Our implementation uses the Wilson method at the 95% confidence level.[51]

Prediction and Uncertainty Sampling

Given the vector of accuracies for each labeling function, $a$, and the vector $l$ for a given article, the prediction $p \in [0, 1]$ is:

$$p = \frac{1}{1 + e^{-x}}, where\ x = \sum_{i}(2l_i - 1) \times \log\frac{a_i}{1 - a_i}$$

We recalibrate the predictions of the generative model using the mean and standard deviations of the articles manually annotated positively and negatively.

In a human-in-the-loop approach, the automated model provides both predictions as well as prioritizing instances for manual annotation [33]. We prioritize articles for manual annotation by identifying articles whose predictions are uncertain. We use two approaches for uncertainty sampling, specifically distance to threshold and prediction variation. Under distance to threshold, instances with predictions closer to the decision boundary have higher uncertainty; given a prediction $p$ and a threshold $t$, the distance to the threshold is $dist = abs(p - t)$. We support uncertainty sampling via variation by running the data programming inference multiple times, masking 50% of the labeling functions during each run, then calculating the inter-quartile range of the predictions for each instance ($iqr$). Our final selection criterion combines the distance to the threshold and variation calculations, choosing the unlabeled instances that simultaneously minimize $dist$ and maximize $iqr$.

Description of Labeling Functions

Data programming with triplet methods requires the assumption that the labeling functions are conditionally independent given the class.[49] We use eight types of labeling functions, chosen for providing complementary views. Table 1 describes the labeling functions briefly; they are fully described in the Supplemental Materials. Several of the labeling function types (LitSuggest, MeSH headings, entity annotations) require training data. Manually annotated data is split into four parts: one part is reserved for evaluation. The remainder are used to train three independent labeling functions for each labeling function type.

Table 1. Labeling functions used to identify articles relevant to Long Covid. Complete descriptions of each labeling function are provided in the Supplemental materials.

| Name | Description | Requires Training Data |
|---|---|---|

| Long Covid grammar | A purpose-built grammar-based named entity recognition system to identify mentions of Long Covid. Uses the full text, if available | No |
| LitSuggest, trained by query | Predictions from the LitSuggest web-based literature curation tool,[52] trained using a query for positives and random articles for negatives | No |
| LitSuggest, trained on annotations | Predictions from the LitSuggest web-based literature curation tool,[52] trained using the annotated training data | Yes |
| PubTator entity annotations | Entities from the PubTator annotation system,[44] using the full text, if available | Yes |
| MeSH headings | Medical Subject Headings indexed by the NLM indexing team, if available | Yes |
| CoronaCentral Long-Haul Topic | Articles from the CoronaCentral portal,[37] annotated with the Long Haul topic and one or more mentions of SARS-CoV-2 | No |
| CoronaCentral Entity Annotations | Entity annotations provided by the CoronaCentral portal | Yes |
| Bias | Labels all articles as probably not relevant | No |

## Data and Code Availability

The Long Covid collection is publicly available at the LitCovid portal: https://www.ncbi.nlm.nih.gov/research/coronavirus/docsum?filters=e_condition.LongCovid. The classification and mentions for each article are also publicly available: https://ftp.ncbi.nlm.nih.gov/pub/lu/LongCovid/. The code used in this study is available for review at https://ftp.ncbi.nlm.nih.gov/pub/lu/LongCovid/.SourceCode. This will be shared publicly through GitHub upon acceptance.

## Acknowledgements

This research was supported by the Intramural Research Program of the National Library of Medicine, National Institutes of Health.

## Author contributions

Conceptualization: RL and ZL. Overall methodology, software, and analysis: RL. Supervision: ZL. Annotation guidelines: RL, RI, JW, AA, and QC. Data curation and validation: RL and JW. PDC analysis: RI. LitCovid resources and software: AA and QC. LitSuggest software: AA. Drafted initial manuscript: RL. All authors reviewed, edited, and approved of the final manuscript.

## Declaration of interests

The authors declare no competing interests.

# Supplemental Material

This supplementary material is divided into four sections. The first section describes the annotation guidelines for determining if an article should be labeled as relevant or not relevant to Long Covid. The second section provides detailed descriptions of the labeling functions used in our framework. The third section details the PubMed queries used for comparison. The final section describes additional experimental results.

## Annotation Guidelines

Following initial CDC guidance on post-COVID-19 conditions, an article is relevant for Long Covid if it contains useful discussion of symptoms that persist 4 weeks (or more) after COVID-19 symptoms start.[1] Relevant articles must therefore (1) assert that the symptoms are probably caused by COVID-19 and (2) discuss those symptoms meaningfully over a timeframe of 4 weeks or more after the start of COVID-19 symptoms.

*Relevant:*
- Any symptom presented as likely caused by COVID-19 – at least in part – is relevant.
    - Lack of a positive COVID test or the potential for other causes (including treatments) do not preclude the article from being relevant.
    - Caused in this sense is counterfactual: if the patient would not have the symptoms if they had not contracted COVID-19, then COVID-19 caused it.
    - These are often mentioned as sequelae, as a secondary disease or as being "after" COVID-19, though none of these establish timing.
- The symptoms may be any type of disease, syndrome, disability, or abnormal finding. These include secondary diseases, psychological effects, sub-clinical abnormalities, and reduction in quality of life.
- New symptoms that appear during the recovery period may still be relevant if they meet all other criteria.
- Articles that do not identify the timeframe but use phrases like "survivor," "recovered" or "long-term" are relevant.
    - Note that "post" and "after" often refer to immediately after. These should be checked carefully to determine if they meet the 4-week criteria.
- Articles that do not identify a timeframe but contain useful discussion of conditions that are long-term, chronic, or permanent (other than death) are relevant.
- Articles describing persistent viral positivity or viral reactivation are relevant.
    - Note that articles on reinfections are not relevant. This may require careful consideration, but in general if the authors present it as probably persistent or a reactivation it is relevant.
- Articles that look for persistent symptoms in the right timeframe in a cohort of COVID-19 survivors are still relevant, even if they conclude that there are none, that they are rare or that they are only mild.
    - These are often discussed as outcomes in a particular timeframe.
- Articles do not need to mention Long Covid, post-acute sequelae of SARS-CoV-2 or any other synonym to be relevant.

- Articles that address Long Covid using an animal model are relevant.

*Not relevant:*
- Articles that only consider the acute phase of COVID-19, i.e. up to 4 weeks after symptoms start.
- Articles that only consider viruses besides SARS-CoV-2 and only diseases other than COVID-19 are not relevant.
- Articles on effects of the pandemic, such as physician burnout or the difficulty of getting treatment.
- Articles on the effects of pandemic lockdowns, such as economic disparities.
- Articles on reinfection are not relevant; these may need to be considered carefully since it is not always clear whether there was a reactivation or reinfection.
- Articles describing adverse effects of SARS-CoV-2 vaccines are not relevant.
- Articles describing an unusual presentation of COVID-19 are a common false positive: while they describe symptoms caused by COVID-19 that would be relevant, those symptoms are usually the initial manifestation of the condition and have resolved by 4 weeks.
- Articles that do not contain a useful or meaningful discussion of the long-term effects of COVID-19 in passing are not relevant. Mentioning sequelae in passing is insufficient.
    - E.g. Articles that do mention Long Covid (e.g. "long-term effects of SARS-CoV-2 infection") but only to say that those effects are unknown or should be identified are not relevant.
    - Articles that mention long-term conditions that are sequelae of COVID-19, but do not discuss them are not relevant.

## Labeling Function Descriptions

In data programming, the practitioner creates a set of labeling functions to give labels to input instances. These labeling functions may take many forms, including rule-based patterns, dictionary lookups and noisy supervised classifiers. Importantly, data programming expects these labeling functions to disagree and to be unable to provide labels for some instances. The labeling functions are applied to a large amount of unlabeled data, and the agreement rates between the labeling functions are then used to infer their respective accuracies. These accuracies then form the parameters of a generative model that can be used to label new data points, applying a small amount of knowledge – in the form of labeling functions – to label a large amount of data with accuracy.

*Labeling Function: Long Covid grammar*

We created a grammar-based method to identify mentions of Long Covid by extending our previous work identifying variations in terms for COVID-19 and SARS-CoV-2.[2] We searched lexicons containing medical terms for Long Covid, identifying equivalent concepts in MeSH (Medical Subject Headings),[3] ICD-10 and Wikidata.[4] Supplementary Table 1 summarizes the Long Covid concepts found and their terms. Equivalent concepts were identified in other vocabularies, though no additional terms. We defined the initial grammar for identifying Long Covid terms using this list and iteratively updated the grammar during the annotation process. We apply the Long Covid grammar to the title and abstract of each article and convert the grammar output to a binary labeling function by returning 1 for articles that contain at least one Long Covid term and 0 for all others. If the full text is available for text mining, we also applied the Long Covid grammar to the full text, again returning 1 for articles that contain at least one Long Covid term and 0 for articles that do not. For articles where the full text is not available, we return 0.5 to cause that instance to abstain from the agreement calculations.

*Supplementary Table 1.* Long covid concepts in three resources: Medical Subject Headings (MeSH), International Classification of Diseases, version 10 (ICD-10), and Wikidata. Our grammar-based named entity recognition system for Long Covid was initially defined to recognize all names listed and was iteratively updated as articles were annotated.

| Resource | MeSH | ICD-10 | Wikidata |
| --- | --- | --- | --- |
| Identifier | C000711409 | U09.9 | Q100732653 |
| Preferred concept name | post-acute COVID-19 syndrome | Post COVID-19 condition | Long COVID |
| Alternate concept names | <ul><li>chronic COVID syndrome</li><li>long COVID</li><li>long haul COVID</li><li>long hauler COVID</li><li>long-COVID</li><li>long-haul COVID</li><li>post-acute COVID syndrome</li><li>post-acute COVID19 syndrome</li><li>post-acute sequelae of SARS-CoV-2 infection</li></ul> | <ul><li>Post-acute sequela of COVID-19</li><li>long COVID-19</li><li>COVID-19 sequelae</li><li>Long COVID</li><li>SARS-CoV-2 sequelae</li><li>Sequelae of COVID-19</li><li>Sequelae of SARS-CoV-2</li><li>Post-COVID Syndrome</li></ul> | <ul><li>Post-COVID-19 syndrome</li><li>Post-COVID syndrome</li><li>Post-Acute Sequelae of SARS-CoV-2 infection</li><li>Long-haul COVID</li><li>Chronic COVID Syndrome</li><li>Post-Acute Sequelae of Severe acute respiratory syndrome coronavirus 2</li><li>Post-Coronavirus Disease syndrome</li><li>Post-Coronavirus Disease-2019 syndrome</li><li>Long-haul Coronavirus Disease</li><li>Chronic Coronavirus Disease Syndrome</li></ul> |

*Labeling Function: LitSuggest*

The LitSuggest web-based literature curation tool provides an easy-to-use interface for annotating PubMed articles,[5] training relevance prediction models and performing predictions. LitSuggest is based on supervised machine learning and therefore requires training data. However, it can provide negative examples by randomly selecting articles not in the positive set.

We used LitSuggest in two ways. We trained LitSuggest using the articles that match a PubMed query as the positive examples and a randomly selected set of articles as the negative set. The PubMed query we use is as follows:

```
"long covid" OR ("post acute sequelae" AND ("SARS-CoV-2" OR "COVID-19"))
```

This has the effect of causing articles that contain vocabulary similar to articles that clearly mention Long Covid to be classified as positive and others to be classified as negative. We then use this model to give a prediction in the range [0, 1] for every article in LitCovid and use these results directly as a labeling function without binarization.

We also trained LitSuggest on the manually labeled data. As an initial set of articles, we manually queried PubMed for "Long Covid" and "post-acute sequelae of SARS-CoV-2 infection," then annotated the results as relevant or not relevant to Long Covid. To better support identifying uncertainty within the predictions, we break the articles annotated so far into four non-overlapping sets. One set is reserved for evaluation while the remaining three sets are used to train separate LitSuggest models, which provide independent predictions. Each of the three models then provide one labeling function, which abstains from providing a prediction for articles seen during training, i.e., returns 0.5.

*Labeling Function: PubTator entity annotations*

PubTator is a web-based system providing automated annotations of six biomedical entity types: genes & proteins, diseases, chemicals, mutations, species, and cell lines.[6] While PubTator does not identify mentions of Long Covid, many entities are correlated with relevance to Long Covid, such as the species SARS-CoV-2 (NCBI Taxonomy 2697049), the symptom Fatigue (MeSH D005221) and the chemical Steroids (MeSH D013256). We use the PubTator API to download the annotations for all articles, then train a logistic regression classifier using the number of times that each entity appears as features. We use the same training data splits as with LitSuggest and create three separate models which provide independent predictions. As with LitSuggest, each of the three models then provide one labeling function, which abstains (returns 0.5) for articles seen during training. If the full text of the article is available for text mining, we also download the PubTator results on the full text. We create an additional three labeling functions with this input using a similar approach as the PubTator results on the title and abstract, with two primary differences. The first difference is that the features represent both the entity identified and the section where the entity appears, such as introduction, methods, or results. Second, if the full text is not available, the full text PubTator entity labeling functions abstain from providing a label (i.e. return 0.5).

*Labeling Function: MeSH headings*

Many articles in PubMed are also indexed with MeSH terms (Medical Subject Headings). MeSH terms are added manually by the indexing team at the National Library of Medicine. As of mid-November 2021, there are very few PubMed articles indexed with the MeSH term for Long Covid. However, many terms are correlated, such as COVID-19 (D000086382) and Survivors (D017741). We use the NBCI e-utilities to download both MeSH terms and publication types (e.g. Journal article) for all articles.[7] We then train three independent models using the same approach as for the PubTator entity annotations, though each feature will have a value of 0 or 1 since MeSH terms appear at most once.

*Labeling Function: CoronaCentral Long Haul Topic*

CoronaCentral is a literature portal for coronaviruses, including SARS and MERS as well as SARS-CoV-2.[8] It provides annotations of both article topics and various entity types. We use CoronaCentral to create a binary labeling function by considering articles annotated with the Long Haul topic and at least one annotation of SARS-CoV-2 to be relevant for Long Covid. All other articles are considered not relevant.

*Labeling Function: CoronaCentral Entity Annotations*

CoronaCentral provides automated entity annotations for 13 entity types using WikiData identifiers. These entities include species (e.g. SARS-CoV-2, Q82069695), genes and proteins (ACE-2, Q14875321), disorders (diabetes, Q12206) and others, such as social distancing (Q30314010). Like the PubTator entity annotations, many of these are correlated with relevance to Long Covid. We therefore create three independent predictors, using the same approach as for the PubTator entity annotations.

*Labeling Function: Bias*

Finally, the bias labeling function predicts that every article – regardless of content – is probably negative. We use the number of positive and negative predictions in the previous round to determine the probability of any given article being positive, then use this value as the prediction for all articles.

## Evaluation Details

### Comparison Collections

The predefined query for Long COVID in PubMed Clinical Queries is as follows:[9]

```
"post acute sequelae of Sars-CoV-2" OR ("PASC" AND ("COVID-19" OR "Sars-CoV-2")) OR "post acute sequelae of COVID" OR "COVID-19 sequelae" OR "long haul covid" OR "covid long haul*" OR "long covid" OR "long term covid" OR "chronic covid syndrome" OR "post covid syndrome" OR "post COVID-19 neurological syndrome" OR "post-acute COVID-19 syndrome" [Supplementary Concept] OR "COVID-19 post-intensive care syndrome" [Supplementary Concept]
```

The MeSH query we created to identify Long Covid articles is as follows:

```
(post-acute COVID-19 syndrome[MeSH Terms]) OR (((COVID-19[MeSH Terms]) OR (SARS-CoV-2[MeSH Terms])) AND ((Quality of Life[MeSH Terms]) OR (Survivors[MeSH Terms]) OR (Recovery of Function[MeSH Terms]) OR (Aftercare[MeSH Terms]) OR (Rehabilitation[MeSH Terms])))
```

## Additional Results

### Contribution of Labeling Functions: Ablation Study

We analyzed the contribution of each labeling function to the overall predictions using an ablation study. We calculated the ROC AUC of the system after ablating each of the labeling functions, and the change in the AUC from the ROC AUC of the full system (0.8483). Removing the LitSuggest labeling functions causes a drop in the ROC AUC of 0.1059 to 0.7424. The change in ROC AUC when removing any of the other labeling functions is small or negligable (data not shown). By creating an ensemble from many data sources, our framework allows the overall model to be robust to removing individual sources.

### Comparison of Topics between LitCovid and the Long Covid Collection

The LitCovid portal annotates all articles with abstracts with the following 8 topics: General, Mechanism, Transmission, Diagnosis, Treatment, Prevention, Case Report, and Forecasting. Topics are annotated automatically, with low-confidence articles subsequently reviewed manually. Any number of topics may be annotated, except for General and Case Report, which – if chosen – are not annotated with any other topic. We compared the prediction rates for each topic within LitCovid and the Long Covid collection. We report the results in Supplementary Figure 1a, using the annotation rate per 1000 articles, and eliding the Transmission and Forecasting topics, which are not applicable. Note that the difference is statistically significant for all topics (p < 0.01, Fisher's exact test). In Supplementary Figure 1b we plot the number of articles over time for each of the four largest topics in the Long Covid collection.

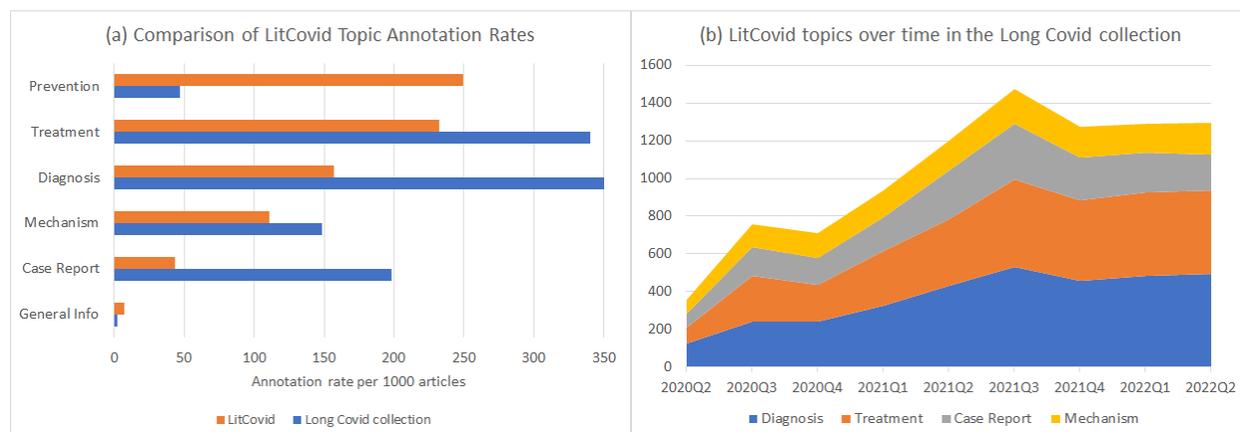

*Supplementary Figure 1*. Analysis of the LitCovid topics in the Long Covid collection. (a) Comparison of the annotation rate per thousand articles for six topics between LitCovid and the Long Covid Collection. Topics Transmission and Epidemic Forecasting are not shown. All comparisons show a statistically significant difference ($p < 0.01$, Fisher's exact test). (b) Number of articles over time for each of the four most common LitCovid topics in the Long Covid collection.

The annotation rate in the Long Covid Collection for the Case Report topic is nearly five times the rate in LitCovid, the rate for the Diagnosis topic is more than double, and both the Mechanism and Treatment topics are also higher. Only the annotation rates for General Info and Prevention topics are lower in the Long Covid Collection. The disease-causing mechanism for COVID-19 is better understood than the mechanism for Long Covid, which drives an increased focus in Long Covid on not only mechanism, but also diagnosis and treatment. Considering the number of articles over time, we see that both the Diagnosis and Treatment topics are increasing more over time than the Mechanism or Case Report topics. This suggests that articles on Long Covid to date have a stronger clinical focus than articles on COVID-19 in general.